\pdfoutput=1
\documentclass[sigconf]{acmart}
\usepackage{hyperref}
\hypersetup{
    colorlinks=true,
    linkcolor=black,
    urlcolor=black,
    citecolor=black,
    breaklinks=true 
}

\def\equationautorefname~#1\null{Equation~(#1)\null}
\usepackage{breakurl}
\usepackage{color}
\usepackage{bm}
\usepackage{xspace}
\usepackage{algorithmic}
\usepackage{algorithm}
\usepackage{mdwlist}    
\usepackage{paralist} 

\usepackage{utfsym}

\usepackage{adjustbox}
\usepackage{array}
\usepackage{booktabs}
\usepackage{multirow}

\usepackage{array,graphicx}
\usepackage{booktabs}
\usepackage{pifont}
\usepackage{color}

\usepackage{colortbl}
\definecolor{lightgray}{gray}{0.85}
\usepackage{multirow} 

\usepackage{fancybox} 
\usepackage{amsmath}
\usepackage{amsfonts}

\newcommand{\vswd}{\vspace{0.0em}}
\newcommand{\bit}{\vspace{-0em}\begin{itemize}}
\newcommand{\eit}{\end{itemize}\vspace{-0.2em}}
\newcommand{\ben}{\vspace{-0em}\begin{enumerate}}
\newcommand{\een}{\end{enumerate}\vspace{-0.2em}}
\newcommand{\bea}{\vspace{-0em}\begin{eqnarray}}
\newcommand{\eea}{\end{eqnarray}\vspace{-0.0em}}
\newcommand{\beq}{\vspace{-0.0em}\begin{equation}}
\newcommand{\eeq}{\end{equation}\vspace{-0.0em}}

\renewcommand{\bit}{\vswd\begin{compactitem}}
\renewcommand{\eit}{\end{compactitem}\vswd}
\renewcommand{\ben}{\vswd\begin{compactenum}}
\renewcommand{\een}{\end{compactenum}\vswd}

\newcommand{\argmax}{\mathop{\rm arg~max}\limits}

\newcommand{\score}[1]{score(#1)}
\newcommand{\norm}{norm}

\newcommand{\hide}[1]{}

\newcommand{\mypara}[1]{\vspace{1.00em}\noindent\textbf{#1.}}

\newcommand{\myparaitemize}[1]{\noindent{\textbf{#1.}}}

\newtheorem{problem}{Problem}
\newtheorem{definition}{Definition}

\newcommand{\attribute}{attribute\xspace}
\newcommand{\attributes}{attributes\xspace}

\newcommand{\units}{units\xspace}

\newcommand{\topic}{component\xspace}
\newcommand{\topics}{components\xspace}

\newcommand{\method}{\textsc{CyberCScope}\xspace}
 
\newcommand{\tensor}{\mathcal{X}}
\newcommand{\tensorC}{\mathcal{X}^c}
\newcommand{\Matt}{\mathbf{A}}
\newcommand{\Mtime}{\mathbf{B}}

\newcommand{\Patt}{\mathbf{\hat{A}}}
\newcommand{\Ptime}{\mathbf{\hat{B}}}

\newcommand{\regime}{\theta}
\newcommand{\regimeset}{\Theta}
\newcommand{\regimeassignment}{\mathcal{S}}
\newcommand{\regimeassign}{s}
\newcommand{\cand}{\mathcal{C}}

\newcommand{\costM}[1]{<#1>}
\newcommand{\costC}[2]{<#1|#2>}
\newcommand{\costT}[2]{<#1;#2>}

\newcommand{\nunits}{U}
\newcommand{\nmode}{M_1}
\newcommand{\cmode}{M_2}
\newcommand{\ntopic}{K}
\newcommand{\duration}{T}
\newcommand{\nregime}{R}
\newcommand{\nshiftp}{G}

\newcommand{\lmode}{m_1}
\newcommand{\lcmode}{m_2}
\newcommand{\ltopic}{k}
\newcommand{\ltime}{t}
\newcommand{\lregime}{r}
\newcommand{\lshiftp}{g}

\newcommand{\mathN}{\mathbb{N}}
\newcommand{\mathR}{\mathbb{R}}

\newcommand{\segmentset}{\mathcal{S}}

\newcommand{\cisev}{\textit{CI'17}\xspace}
\newcommand{\ciei}{\textit{CCI'18}\xspace}

\copyrightyear{2025}
\acmYear{2025}
\setcopyright{acmlicensed}\acmConference[WWW Companion '25]{Companion Proceedings of the ACM Web Conference 2025}{April 28-May 2, 2025}{Sydney, NSW, Australia}
\acmBooktitle{Companion Proceedings of the ACM Web Conference 2025 (WWW Companion '25), April 28-May 2, 2025, Sydney, NSW, Australia}
\acmDOI{10.1145/3701716.3715476}
\acmISBN{979-8-4007-1331-6/25/04}

\author{Kota Nakamura}
\affiliation{%
  \institution{SANKEN, Osaka University}
  \state{Osaka}
  \country{Japan}
}
\email{kota88@sanken.osaka-u.ac.jp}
\author{Koki Kawabata}
\affiliation{%
  \institution{SANKEN, Osaka University}
  \state{Osaka}
  \country{Japan}
}
\email{koki@sanken.osaka-u.ac.jp}
\author{Shungo Tanaka}
\affiliation{%
  \institution{SANKEN, Osaka University}
  \state{Osaka}
  \country{Japan}
}
\email{tanaka.shungo88@sanken.osaka-u.ac.jp}
\author{Yasuko Matsubara}
\affiliation{%
  \institution{SANKEN, Osaka University}
  \state{Osaka}
  \country{Japan}
}
\email{yasuko@sanken.osaka-u.ac.jp}
\author{Yasushi Sakurai}
\affiliation{%
  \institution{SANKEN, Osaka University}
  \state{Osaka}
  \country{Japan}
}
\email{yasushi@sanken.osaka-u.ac.jp}

\begin{document}
\title{\method: 
Mining Skewed Tensor Streams 
and \\
Online Anomaly Detection 
in Cybersecurity Systems}
\begin{abstract}
    Cybersecurity systems
are continuously producing a huge number of 
time-stamped events 
in the form of high-order tensors, such as
%
\{{\it
count; time, port, flow duration, packet size, \ldots} \},
and so 
how can we detect anomalies/intrusions in real time?
How can we identify multiple types of intrusions and capture 
their characteristic behaviors?
The tensor data consists of categorical and continuous attributes
and the data distributions of continuous attributes typically 
exhibit skew.
These data properties require handling skewed infinite and finite dimensional spaces simultaneously.
In this paper, we propose 
a novel streaming method, namely \method.
The method effectively decomposes incoming tensors into major trends 
while explicitly distinguishing between categorical and skewed continuous attributes.
To our knowledge, it is the first to compute 
hybrid skewed infinite and finite dimensional decomposition.
Based on this decomposition, 
it streamingly finds distinct time-evolving patterns, enabling 
the detection of multiple types of anomalies.
Extensive experiments on 
large-scale 
real datasets demonstrate that 
\method 
detects various intrusions with higher accuracy than
state-of-the-art baselines while 
providing meaningful summaries for the intrusions 
that occur in practice.
\end{abstract}



\maketitle
\section{Introduction}
    \label{010intro}
    Cybersecurity systems monitor web-scale data streams 
that are increasingly larger in size and faster in transaction speed.
Streaming anomaly detection 
aims to efficiently analyze these data streams and accurately identify the sudden appearance of anomalies (e.g, intrusions) in real time.

Recent systems enable us to access a massive volume and 
variety of data streams, 
represented as high-order tensor streams
consisting of time-stamped events with multiple attributes,
such as \textit{(time, port, flow duration, packet size, \ldots)}.
Handling the high-dimensional data is particularly challenging for traditional anomaly detection algorithms, such as One-Class SVM, which tend to perform poorly due to the curse of dimensionality.
Effective methods for analyzing 
tensor streams (or multi-aspect data)
has been extensively studied
\cite{shin2017densealert, manzoor2018xstream,
kawabata2020non,bhatia2021mstream,DBLP:conf/www/0001JSKH22}.
CubeScope \cite{nakamura2023fast} 
can detect anomalies/intrusions
with interpretable summaries of tensor streams,
such as distinct time-evolving patterns
and major trends in attributes. 

\begin{figure}
    \hspace{-2em}
    \centering    
    \includegraphics[width=0.88\columnwidth]{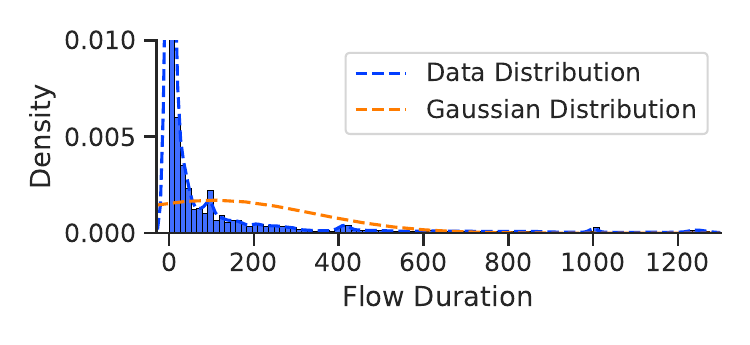}
    \vspace{-1em}
    \caption{Data distribution of 
    a continuous attribute is skewed:
    it exhibits right skewness, deviating from a Gaussian distribution
    based on the empirical mean and variance.}
\label{fig:data_dist}
\end{figure}

However, practical application to cybersecurity systems remains challenging 
due to the following two data properties.
\textit{(a) Tensor data consist of categorical and continuous attributes.}
Let us consider analyzing a collection of time-stamped events 
with two attributes: port and flow duration.
The port can be represented as categorical values, 
resulting in discrete finite dimensional space.
In contrast, the flow duration is continuous numeric data,
requiring an infinite dimensional space to represent all possible values.
Formally, the data become 3rd-order quasitensor 
$\tensor \in \mathN^{\duration \times \nunits \times \infty}$, 
where $\duration$ is time duration and 
$\nunits$ indicates the unique units for port.
Existing tensor-based approaches handle the infinite dimensional space by discretization, 
which ignores the continuous properties.
%
\textit{(b) The data distributions in continuous attributes are skewed.}
Skewed data distributions are ubiquitous in web-centric domains \cite{korn2006modeling}.
For example, 
Figure~\ref{fig:data_dist} illustrates
the data distribution of a continuous attribute, 
specifically flow duration in the \ciei dataset.
The data distribution exhibits right skewness, 
making Gaussian assumption infeasible.
The ideal method should effectively capture such skewed distributions 
and their multi-way relations in a tensor stream.

In this paper, 
we refer to data streams holding the above properties as 
``\textit{skewed tensor streams}'', 
for which we propose 
an efficient and effective mining approach, namely \method.
%
The approach effectively decomposes incoming tensors into major trends while explicitly distinguishing between categorical and skewed continuous attributes.
To our knowledge, this is the first method to compute hybrid skewed infinite and finite dimensional decomposition 
(see \cite{larsen2024tensor} for details on the mathematical concepts).
Building on this decomposition, 
\method streamingly finds distinct time-evolving patterns,
referred to as ``\textit{regimes}''.
Although tensor streams in cybersecurity systems 
may contain 
multiple types of intrusions and newly emerged ones,
regimes enable us to identify the types of anomalies 
and effectively assess the anomalousness of tensors.
Our experimental results on large-scale real datasets show that \method detects various intrusions 
with higher accuracy than state-of-the-art baselines 
while extracting characteristic behaviors 
of the intrusions that occur in practice.
%
%
%

\mypara{Contributions}
The main contributions of our paper are:
\bit
    \item  
        \textit{Modeling Skew}:
        We propose \method based on online probabilistic skewed infinite and finite dimensional (OP-SiFi) decomposition,
        which extracts major trends from tensor streams with skewed continuous attributes.
    \item
        \textit{Algorithm}:
        Our proposed algorithm finds distinct time-evolving patterns 
        (i.e., regimes), 
        which enable us to 
        identify the multiple types of anomalies 
        with their characteristic behaviors.
    \item  
        \textit{Effectiveness}:
        Our experimental results demonstrate that 
        \method outperforms state-of-the-art baselines 
        on large-scale real-world datasets
        while providing an interpretable summary of
        skewed tensor streams in real time.
\eit
\myparaitemize{Reproducibility}
Our source code and datasets are available at~\cite{WEBSITE}.

\section{Proposed Method}
    \label{030model}
    In this section, we describe the tensor streams that 
we want to analyze, 
define the formal problem of streaming anomaly detection, 
and present our method.

Let us consider continuous monitoring of time-stamped events
with $\nmode$ categorical \attributes (e.g., port) and 
$\cmode$ continuous \attributes (e.g., flow duration).
The data takes the form of a $(1+\nmode+\cmode)$-th order 
tensor stream $\tensor$.
$\duration$ indicates the most recent time.
For the $\lmode$-th categorical attribute,
we assume a discrete finite dimensional space $\nunits_{\lmode}$.
For the $\lcmode$-th continuous attribute,
we assume an infinite-dimensional space.
For example,
when monitoring time-stamped events with one categorical attribute 
and one continuous attribute, we handle a 3rd-order tensor stream
$\tensor \in \mathN^{\duration \times \nunits_1 \times \infty}$
\footnote{
In this paper, we use $\infty$
to denote the entire space of positive real numbers.}.

At every time point $\duration$ 
that is arrived at with a non-overlapping time interval $\tau \ll \duration$,
we can obtain the current tensor $\tensorC$ 
as the partial tensor of $\tensor$.
In the case of the aforementioned third-order tensor stream,
we continuously obtain a current tensor $\tensorC \in 
\mathN^{\tau \times \nunits_1 \times \infty}$.

As discussed in the introduction,
continuous attributes in cybersecurity systems have skewed data distributions. Thus, we assume that the continuous attributes in the tensor stream $\tensor$ are skewed, referring to $\tensor$ as a \textit{skewed tensor stream}.

Our goal is to detect group anomalies
\cite{bhatia2021mstream},
which are sudden appearances of suspicious similar events
intended to threaten victims, such as the DoS attack,
while their individual activities are small and thus overlooked.
So, how efficiently can we evaluate
anomalousness of the current tensor $\tensorC$ while monitoring the entire tensor stream $\tensor$?
To achieve the goal,
we estimate a compact description $\cand$ of $\tensor$
and define our anomalousness measure 
as
a distance between $\cand$ and arriving $\tensorC$.
An ideal $\cand$ should well capture normal behavior
based on skewed infinite-/finite-dimensional spaces.
It should also be capable of multiple temporal patterns (i.e., regimes) to be aware of multiple types of group anomalies that arise over time.

Consequently,
we define our problem as follows.
\begin{problem}
\label{pro:interanom}
\textbf{Given}
a current tensor $\tensorC$ 
as a partial tensor of a skewed tensor stream $\tensor$,
\bit
\item 
\textbf{Maintain} a compact description $\cand$ for the entire stream $\tensor$,
\item 
\textbf{Report} an anomaly score for the current tensor $\tensorC$,
\eit
continuously, as quickly as possible.
\end{problem}%

\subsection{Proposed Solution: \method}
\begin{algorithm}[t]
    \caption{\method $(\tensorC, \cand)$}
    \label{alg:main}
\begin{algorithmic}[1]
    \REQUIRE
        1. Current tensor
        $\tensorC\in\mathN^{
            \tau \times \nunits_1 \times \ldots \times \nunits_{\nmode} \times 
            \footnotesize \prod_{\nmode}^{\cmode}
            \normalsize \infty 
        }$\\
        \hspace{1.55em}
        2. Previous compact description $\cand = \{\nregime, \regimeset, \nshiftp, \regimeassignment \}$\\
    \ENSURE
        1. Updated compact description $\cand'$\\
        \hspace{2.4em}
        2. Anomalousness score $\score{\tensorC}$
    \STATE 
    {\color{blue}/* Section~\ref{section:decomp} */}
    \STATE $\regime_c$ = OP-SiFi decomposition
    ($\tensorC$);
    \STATE 
    {\color{blue}/* Section~\ref{section:compress} */}
    \STATE $\cand', \score{\tensorC}$
    = MDL-based model compression 
    ($\regime_c, \tensorC, \cand $);
    \STATE {\bf return}  $\cand', \score{\tensorC}$;
\end{algorithmic}
    \normalsize
\end{algorithm}
We now address 
Problem~\ref{pro:interanom} by proposing \method. 
The method continuously extracts major trends 
and their multi-way relations from the current
tensor $\tensorC$.
Then, it updates a compact description $\cand$
and assigns an anomaly score to the current tensor.
Algorithm~\ref{alg:main} shows the overall procedure.

\subsubsection{OP-SiFi Decomposition}
\label{section:decomp}
\begin{figure}
    \centering
    \includegraphics[width=\columnwidth]{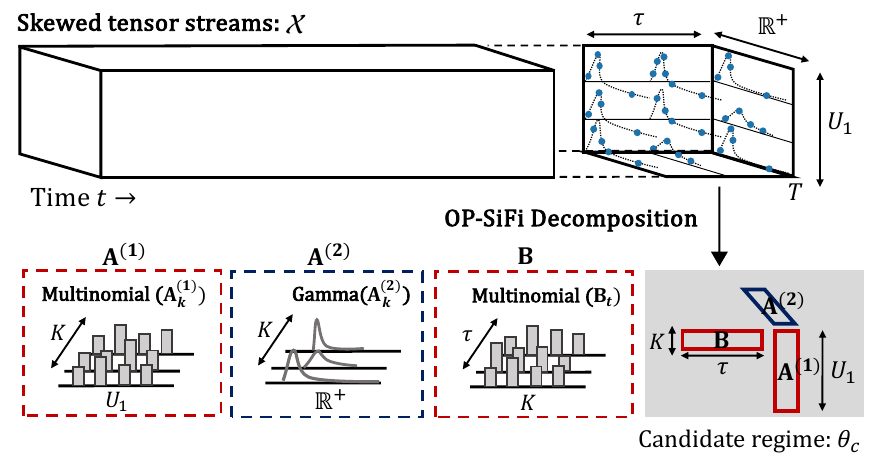}
    \vspace{-2.5em}
    \caption{Overview of OP-SiFi decomposition.}
\label{fig:overview}
\end{figure}

We begin with the simplest case,
where we have only a current tensor $\tensorC$.
Our first step is to decompose
a current tensor $\tensorC$ into major trends 
while distinguishing between categorical and skewed continuous attributes.
We thus propose an 
online probabilistic skewed infinite and finite dimensional
(OP-SiFi) decomposition,
illustrated in Figure~\ref{fig:overview}.
Specifically, 
we assume that there are $\ntopic$ major trends
behind the event collections 
and refer to such trends as \textit{\topic}.
The $\ltopic$-th \topic is characterized
by probability distributions 
in terms of $\nmode$ categorical attributes, $\cmode$ skewed continuous attributes, and time:
\bit
\item
$ \Matt^{(\lmode)}_{\ltopic} \in \mathR^{\nunits_{\lmode}}$:
Multinomial distribution 
over $\nunits_{\lmode}$ \units of the attribute $\lmode$ for
the \topic $\ltopic$.
\item
$ \Matt^{(\nmode+\lcmode)}_{\ltopic} \in 
\mathR_{\footnotesize\mbox{>0}}^{2}$:
Shape and rate (inverse scale) 
parameters of Gamma distribution for
the \topic $\ltopic$.
Note that the gamma distribution is right-skewed when 
the shape parameter 
$\Matt^{(\nmode+\lcmode)}_{\ltopic,1}$ is small, 
whereas it becomes more symmetrical as the shape parameter increases.

\item
$ \Mtime_{\ltime} \in \mathR^{\ntopic}$:
Multinomial distribution over $\ntopic$ \topics
for the time $\ltime \in \tau$.
\eit
We refer to
$\Matt^{(1)}, \ldots, \Matt^{(\nmode)},
\Matt^{(\nmode+1)}, \ldots, \Matt^{(\nmode+\cmode)}$,
and $\Mtime$ as \topic matrices.
The generative process can be described 
as follows:
\begin{center}
  \fbox{
  \hspace{-2em}
  \begin{minipage}{1.0\columnwidth}
\small
\begin{itemize}
\setlength{\parskip}{0cm}\setlength{\itemsep}{0.1cm}
\item 
  For each \topic $\ltopic=1, \ldots, \ntopic$: 
  \begin{itemize}
  \setlength{\parskip}{0cm}\setlength{\itemsep}{0.1cm}
  \renewcommand{\labelenumi}{(\alph{enumi})}
    \item
    For each categorical attribute $\lmode=1, \ldots, \nmode$: 
    \begin{itemize}
    \setlength{\parskip}{0cm}\setlength{\itemsep}{0.1cm}
    \renewcommand{\labelenumi}{(\alph{enumi})}
      \item 
       $ \Matt^{(\lmode)}_{\ltopic} \sim \textrm{Dirichlet}
       (\Patt^{(\lmode)}_{\ltopic})$
    \end{itemize}
  \setlength{\parskip}{0cm}\setlength{\itemsep}{0.1cm}
  \renewcommand{\labelenumi}{(\alph{enumi})}
    \item
    For each continuous attribute $\lcmode=1, \ldots, \cmode$: 
    \begin{itemize}
    \setlength{\parskip}{0cm}\setlength{\itemsep}{0.1cm}
    \renewcommand{\labelenumi}{(\alph{enumi})}
      \item 
       $ \Matt^{(\nmode+\lcmode)}_{\ltopic,2} \sim 
       \textrm{Gamma}(\Patt^{(\nmode+\lcmode)}_{\ltopic})$
    {\color{blue}
    // Rate parameter
    }
    \item 
       $ \Matt^{(\nmode+\lcmode)}_{\ltopic,1} = 
       \mathcal{F}(\Matt^{(\nmode+\lcmode)}_{\ltopic,2},
       \Patt^{(\nmode+\lcmode)}_{\ltopic})$
    {\color{blue}
    // Shape parameter
    }
    \end{itemize}
  \end{itemize}
\item
  For each time $\ltime=1, \ldots, \tau$:
  \begin{itemize}
  \setlength{\parskip}{0cm}\setlength{\itemsep}{0.1cm}
  \renewcommand{\labelenumi}{(\alph{enumi})}
    \item
      $\Mtime_{\ltime} 
      \sim 
      \textrm{Dirichlet}(\Ptime_{\ltime})$
    \item
      For each entry $j=1, \ldots, N_{\ltime}$: 
      \begin{itemize}
      \setlength{\parskip}{0cm}\setlength{\itemsep}{0.1cm}
      \renewcommand{\labelenumi}{(\alph{enumi})}
        \item
        $z_{\ltime,j} \sim \textrm{Multinomial}(\Mtime_{\ltime})$
        {\color{blue}
        // Draw a latent \topic $z_{\ltime,j}$ 
        }
        \item 
        For each categorical \attribute $\lmode=1, \ldots, \nmode$: 
        \begin{itemize}
        \setlength{\parskip}{0cm}\setlength{\itemsep}{0.1cm}
        \renewcommand{\labelenumi}{(\alph{enumi})}
          \item
          $e^{(\lmode)}_{\ltime,j} \sim \textrm{Multinomial}(\Matt^{(\lmode)}_{z_{\ltime,j}})$
        \end{itemize}
        For each continuous \attribute $\lcmode=1, \ldots, \cmode$: 
        \begin{itemize}
          \item
          $e^{(\nmode+\lcmode)}_{\ltime,j} \sim \textrm{Gamma}(\Matt^{(\nmode+\lcmode)}_{z_{\ltime,j}})$
        \end{itemize}
      \end{itemize}
    \end{itemize}
\end{itemize}
\normalsize
\end{minipage}
}
\end{center}
where 
$N_{\ltime}$ is the total number of events at time $t$, 
and $z_{\ltime,j}$ is the latent \topic .
Each event $e_{\ltime,j}$ is sampled from the \topic -specific probabilistic distributions.
$\Patt^{(\lmode)}_{\ltopic}$,
$\Patt^{(\nmode + \lcmode)}_{\ltopic}$,
and $\Ptime_{\ltime}$
are the previous \topic matrices
at $\duration-\tau$
\footnote{
We employ
$\Patt^{(\lmode)}_{\ltopic}=\Ptime_{\ltime} = 
\frac{1}{\ntopic}$
and $\Patt^{(\nmode + \lcmode)}_{\ltopic} = 1$
at the start of the process.}.
We can incorporate the temporal dependencies 
by applying the previous \topic matrices 
as priors \cite{nakamura2023fast}.
Note that 
the conjugate prior for the Gamma rate parameter 
is a Gamma distribution, 
but no proper conjugate prior exists for the shape parameter.
Therefore, we estimate 
the shape parameter using a function $\mathcal{F}$ based on 
Bayesian learning with unnormalized prior \cite{llera2016bayesian}.
According to the generative process, 
we efficiently estimate the \topic matrices
that best describe $\tensorC$
by employing collapsed Gibbs sampling \cite{porteous2008fast}.

\subsubsection{Compact Description}
We here formally define compact description $\cand$ 
by employing component matrices as the building blocks.
Although the \topic matrices 
concisely describe
the partial tensor $\tensorC$,
they are insufficient 
to represent the whole tensor stream $\tensor$,
which contains
various types of distinct dynamical patterns.
We thus introduce another higher-level architecture.

\begin{definition}[Regime: $\regime$]
\rm{
Let $\regime$ be a regime consisting of the \topic matrices:
$\regime = 
\{ 
\{\Matt^{(\lmode)}\}_{\lmode=1}^{\nmode}, 
\{\Matt^{(\nmode+\lcmode)}\}_{\lcmode=1}^{\cmode}, 
\Mtime \}$
to describe a certain distinct dynamical pattern
with which we can divide and summarize the entire tensor stream into segments.
When there are $\nregime$ regimes, 
a regime set is defined as $\regimeset =\{\regime_\lregime\}_{\lregime=1}^{\nregime}$.
}
\end{definition}

A compact description represents the whole tensor stream $\tensor$ by a combination of regimes.
When there are $\nshiftp$ switching positions, 
the regime assignments are defined as $\regimeassignment =\{\regimeassign_\lshiftp\}_{\lshiftp=1}^{\nshiftp}$,
where $\regimeassign_\lshiftp =(t_s,\lregime)$ is the history of each switching position $t_s$ to the $\lregime$-th regime.
Finally, all the parts for a compact description are follows:
\begin{definition}[Compact description]
\rm{
Let $\cand = \{\nregime, \Theta, \nshiftp, \regimeassignment \}$ 
be 
a compact representation of the whole tensor stream $\tensor$,
    namely,
    \bit  
    \item
    the number of regimes $\nregime$ and the regime set, 
    $ \regimeset =\{\regime_\lregime\}_{\lregime=1}^{\nregime}$, 
    \item
    the number of segments $\nshiftp$ and the assignments, 
    $ \regimeassignment =\{\regimeassign_\lshiftp\}_{\lshiftp=1}^{\nshiftp}$.
    \eit
}
\end{definition}
\subsubsection{MDL-based Model Compression}
\label{section:compress}
Our final goal is 
to continuously update the compact description $\cand$
and report an anomaly score of the current tensor $\tensorC$.
Here, we manage the compact description $\cand$
based on the minimum description length (MDL) principle \cite{grunwald2005advances}.
In short, the principle follows the assumption that 
the more we can compress the data, the more we can learn
about their underlying patterns.
We evaluate the total encoding cost,
which can be used to compress 
the original tensor stream $\tensor$.
Specifically,
we estimate a candidate regime $\regime_c$ that 
describes $\tensorC$ by employing OP-SiFi decomposition
and then choose a regime from 
$\regimeset \cup \{\regime_c\}$
so that the additional encoding cost is minimized.
The additional encoding cost $\costT{\tensorC}{\regime_{*}}$ is written as follows:
\begin{center}
\fbox{
\begin{minipage}{0.95\columnwidth}
\small
\begin{align}
    \costT{\tensorC}{\regime_{*}} 
    &= \Delta\costM{\cand} + \costC{\tensorC}{\regime_{*}},\\
    \label{eqn:cost:delta_cost_model}
    \nonumber
    \Delta \costM{\cand}
    \nonumber
    &= \log^{*}(\nregime+1) - \log^{*}(\nregime)~ + \costM{\regime_*} \\
    &+ \log^{*}(\nshiftp+1) - \log^{*}(\nshiftp)~ + \costM{\regimeassign},
  \end{align}    
\normalsize
\end{minipage}
}
\end{center}
where $\regime_*$ indicates any regime.
$\costC{\tensorC}{\regime_{*}}$ represents data coding cost, 
which is the number of bits needed to describe $\tensorC$ 
by employing the regime $\regime_{*}$, i,e.,
$\costC{\tensorC}{\regime_{*}}= -\log P(\tensorC|\regime_{*})$.
$\costM{\cand}$ is the model coding cost, which 
represents the number of bits required to describe the model (see \cite{nakamura2023fast} for details on each term).
If we need 
to shift another existing regime to represent $\tensorC$,
then $\Delta\costM{\cand} = \log^{*}(\nshiftp+1) - \log^{*}(\nshiftp) + \costM{\regimeassign}$
\footnote{
$\log*$ indicates the number of bits for integers based on universal code length.};
if the description of $\tensorC$ requires new regimes, 
it costs all of the terms in \autoref{eqn:cost:delta_cost_model};
otherwise, $\Delta\costM{\cand} = 0$.
\begin{figure*}
    \centering
    \includegraphics[width=\linewidth]{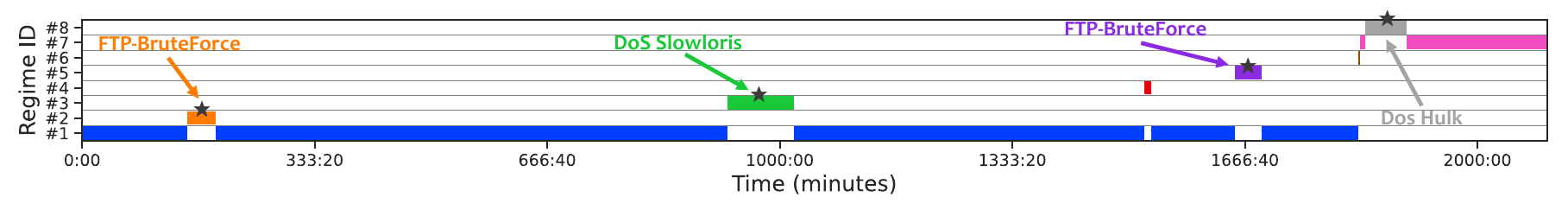}
    \vspace{-3em}
    \caption{Real-time intrusion detection of \method 
    on \ciei dataset:
    the stars indicate intrusions.
    It successfully 
    identified multiple types of intrusions 
    (e.g., \#2 and \#5: FTP-BruteForce, 
    \#3: Dos Slowloris, and \#8: Dos Hulk).}
    \vspace{-1.5em}
\label{fig:anom_effectiveness}
\end{figure*}
\subsubsection{Anomaly Detection}
Finally, we assess the anomalousness of the current tensor $\tensorC$
as follows:
\small
\begin{align}
    &\norm =~ \argmax_{\lregime \in\nregime}|\segmentset^{-1}_{\lregime}|,\\
    &\score{\tensorC}=
    -\log P(\tensorC |\regime_{\norm}),
    \label{eqn:anomaly_score}
\end{align}
\normalsize
where $|\segmentset^{-1}_{\lregime}|$ is
the total segment length of the regime $\regime_\lregime$.
Roughly speaking, we employ the majority regime in the entire tensor stream $\tensor$ as a baseline.
This approach
can adaptively adjust the 
baseline to reflect 
the changes in the nature of the data streams.

\section{Experiments}
    \label{050experiments}
    In this section, 
we evaluate the performance of \method.
We answer the following questions
through the experiments.
\begin{itemize}
\item[(Q1)]
\textit{Effectiveness:}
How successfully does it detect multiple intrusions and 
provides characteristic behaviors of the intrusions?
\item[(Q2)]
\textit{Accuracy:}
How accurately does it achieve streaming anomaly detection?
\item[(Q3)]
\textit{Scalability:}
How does it scale in terms of computational time?
\end{itemize}
\myparaitemize{Datasets}
We use two real datasets,
\cisev \cite{sharafaldin2018toward} 
and \ciei \cite{cicids18}.
These datasets consist of up to $18$ million event logs, 
in which various types of intrusions, such as
brute force attacks and DoS attacks, occur over time.
The attributes for time-stamped events are \textit{(Dst Port, Flow Duration, Total Length of Fwd Packet, Total Length of Bwd Packet, Fwd Header Length, Bwd Header Length, Flow IAT Mean)}, 
forming in $8$th-order 
skewed tensor streams,
where \textit{Dst Port} is the only a categorical attribute.
The study \cite{liu2022error}  
reported errors in these datasets and released improved versions, which we used throughout the experiments.
We set the size of current tensor $\tau$ to
$4$ minutes for the \cisev dataset 
and $30$ seconds for the \ciei dataset,
ensuring that each tensor contains at least one event.

\myparaitemize{Baselines}
Our experiments are evaluated with two state-of-the-art baselines for streaming anomaly detection:
(a) MemStream \cite{DBLP:conf/www/0001JSKH22},
which is a streaming approach using a denoising autoencoder and a memory module.
We set the memory size $N=64$ and 
the threshold for concept drift $\beta=0.01$.
(b) CubeScope \cite{nakamura2023fast},
which is an online factorization method based on probabilistic generative models. The number of components is set to $K=48$.
For \method, we set the number of components to $K=48$. 

\begin{figure}[t]
    \begin{tabular}{cc}
    \begin{minipage}{0.5\columnwidth}
    \centering
    \hspace{-2.5em}
     \includegraphics[width=1\linewidth]{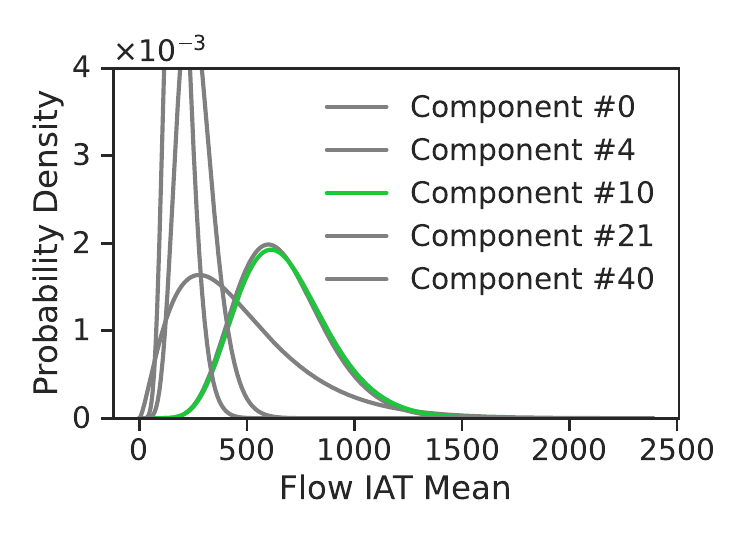}\\
    \vspace{-0.8em}
    \hspace{-1.5em}
     (a) Normal (Regime \#1)
     \end{minipage}
    &
    \begin{minipage}{0.5\columnwidth}
     \centering
     \hspace{-5em}
    \includegraphics[width=1\linewidth]{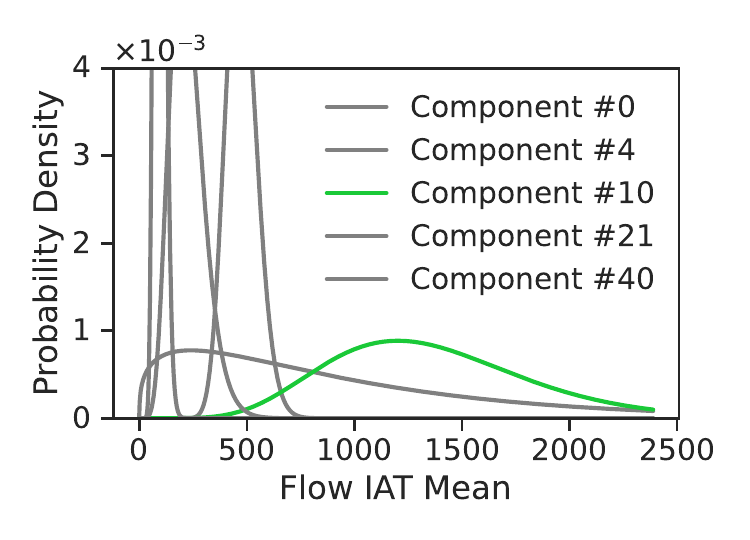}\\
    \vspace{-0.8em}
    \hspace{-4em}
    (b) Dos Slowloris (Regime \#3)
     \end{minipage}
    \end{tabular}
    \vspace{-1em}
    \caption{
    \method captures characteristic behavior of the Dos Slowloris: Component \#10 shifts a larger value.
    }
    \label{fig:shift_slowloris}
    \includegraphics[width=1\columnwidth]{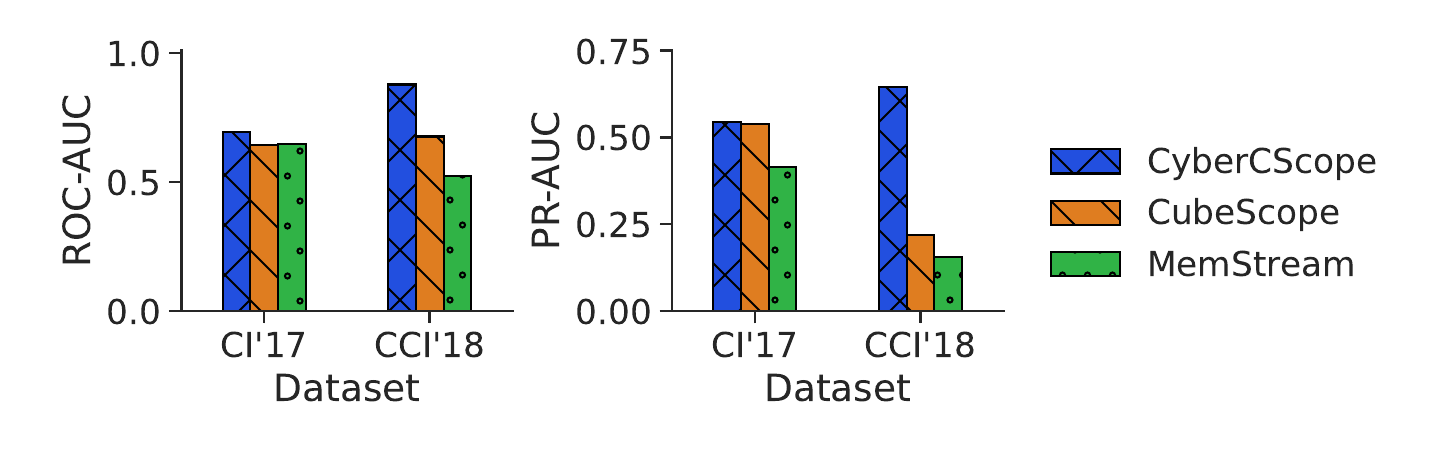}
    \vspace{-3em}
    \caption{ 
    Detection accuracy with respect to 
    ROC-AUC and PR-AUC (higher is better).
    }
    \label{fig:detection_acc}
    \centering
    \hspace{-2em}
    \begin{tabular}{cc}
    \begin{minipage}{0.46\columnwidth}
    \centering
     \includegraphics[width=1\linewidth]{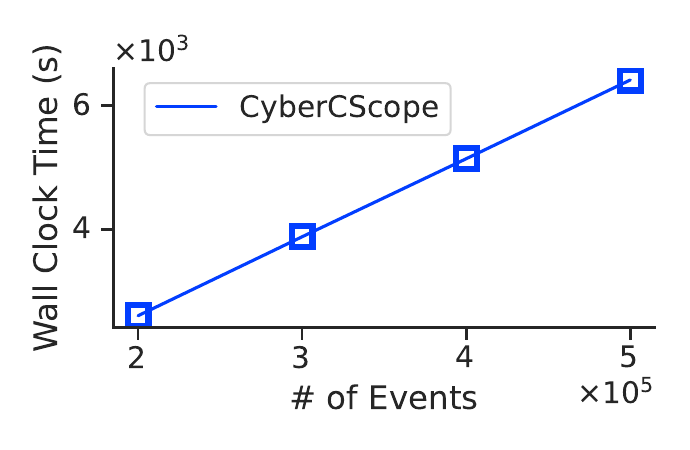}\\
     \end{minipage}
    &
    \begin{minipage}{0.54\columnwidth}
     \centering
         \hspace{-3em}
    \includegraphics[width=1\linewidth]{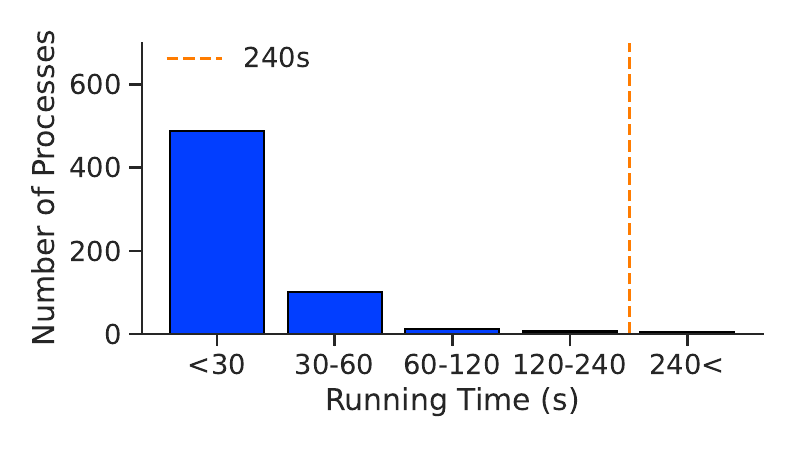}\\
     \end{minipage}
    \end{tabular}
    \vspace{-1.25em}
    \caption{
    Scalability of \method: 
    (left) It scales linearly.
    (right) It processes each tensor in real time.
    }
    \label{fig:saclability}
\end{figure}

\myparaitemize{Q1. Effectiveness}
\label{sec:Effectiveness}
We first demonstrate 
the real-time intrusion detection of \method on the \ciei dataset.
As shown in \autoref{fig:anom_effectiveness},
\method identifies multiple types of intrusions
by detecting regimes.
For example,
Regime~\#2~(orange) and Regime~\#5~(violet) 
correspond to FTP-BruteForce.
Regime~\#8~(gray) is coincided with Dos Hulk.
The DoS Slowloris attack sends requests at long intervals to keep server connections open and exhaust resources.
Figure~\ref{fig:shift_slowloris} shows 
the changes in the top-5 components based on their likelihoods 
when detecting Regime \#3 (Dos Slowloris).
Here, 
we observed intrusion-specific behavior
in the Flow IAT Mean attribute (i.e., mean of inter-arrival time between packet flows):
Component \#10 shifts larger value.
Note that these intrusions occur over time,
making their numbers, durations, and features unknown
\textit{a priori}, whereas the method successfully captures them from data streams.

\myparaitemize{Q2. Accuracy}
\label{subsec:accuracy}
We next evaluate the accuracy of \method
in terms of anomaly detection.
Figure~\ref{fig:detection_acc} shows 
ROC-AUC and PR-AUC for each method, 
where a higher value indicates better detection accuracy.
\method achieves a high detection accuracy for every dataset,
while other methods cannot detect anomalies very well.
The most competitive method, 
CubeScope, captures multi-aspect features in events but handles continuous attributes by discretization,
failing to capture their continuous and skewed properties.

\myparaitemize{Q3. Scalability}
\method is carefully designed to scale linearly 
with the number of events.
The left part of Figure~\ref{fig:saclability} shows 
the computational time of when varying the size of an input tensor
stream, confirming 
the linear scalability of the method. 
The right part of Figure~\ref{fig:saclability} 
shows a frequency distribution of the time 
taken to process each current tensor in the \cisev dataset.
Most processes were completed within four minutes.
This means the method mostly reports
the anomaly scores without delay for the data stream.
\section{Conclusion}
    \label{060conclusions}
    In this paper,
we focused on 
mining skewed tensor streams and 
detecting anomalies 
in cybersecurity systems,
for which we presented \method.
A key part of the method,
OP-SiFi decomposition, captures major trends in
tensor streams over skewed infinite and finite dimensional spaces.
The proposed algorithm detects multiple types of anomalies by 
finding time-evolving patterns.
Through experiments,
\method
detected various intrusions that occur in practice
with higher accuracy than state-of-the-art baselines while
extracting characteristic behaviors of the intrusions 
in real time.

\begin{acks}
This work was partly supported by
JSPS KAKENHI Grant-in-Aid for Scientific Research Number
JP22K17896,
JP24KJ1615,
JST CREST JPMJCR23M3.
\normalsize
\end{acks}
\bibliographystyle{ACM-Reference-Format} 

\end{document}